\documentclass[10pt,twocolumn,letterpaper]{article}

\usepackage{iccv}
\usepackage{times}
\usepackage{epsfig}
\usepackage{graphicx}
\usepackage{amsmath}
\usepackage{amssymb}
\usepackage[accsupp]{axessibility}

\usepackage{graphicx}
\graphicspath{{figure}, {images}, {example}}
\DeclareGraphicsExtensions{.png,.PNG,.jpeg,.JPEG,.jpg,.JPG,.bmp,.BMP}
\usepackage{epsfig} % for figures
\usepackage{subcaption}
\usepackage[export]{adjustbox}
\usepackage{float}
\usepackage[justification=raggedright]{caption}	% makes captions ragged right - thanks to Bryce Lobdell
\usepackage{lscape} % Useful for wide tables or figures.
\usepackage{overpic}

\usepackage[lined,ruled,linesnumbered]{algorithm2e}

\SetAlFnt{\small}
\SetAlCapFnt{\small}
\SetAlCapNameFnt{\small}
\SetAlCapHSkip{0pt}
\IncMargin{-\parindent}

\usepackage{booktabs}  % Publication quality tables
\usepackage{multirow}
\usepackage{paralist}
\usepackage{enumitem}

\usepackage{mathtools}
\usepackage{amsmath}   % Short math guide for LaTeX ftp://ftp.ams.org/pub/tex/doc/amsmath/short-math-guide.pdf
\usepackage{units}
\usepackage{color}

\usepackage{comment}
\usepackage{bbm}
\usepackage{soul}
\usepackage{wrapfig}

\usepackage{array}
\usepackage{tabularx}

\definecolor{purple}{cmyk}{0.45,0.86,0,0}
\definecolor{bleudefrance}{rgb}{0.19, 0.55, 0.91}
\definecolor{darkorange}{rgb}{1, 0.55, 0}
\definecolor{limegreen}{rgb}{0.2, 0.8, 0.2}

\newlength\paramargin
\newlength\figmargin
\newlength\secmargin
\newlength\figcapmargin

\newcommand{\mpage}[2]
{
\begin{minipage}{#1\linewidth}\centering
#2
\end{minipage}
}

\long\def\ignorethis#1{}

\usepackage[pagebackref=true,breaklinks=true,letterpaper=true,colorlinks,bookmarks=false]{hyperref}

\iccvfinalcopy % *** Uncomment this line for the final submission

\def\iccvPaperID{10324} % *** Enter the ICCV Paper ID here

\ificcvfinal\fi

\begin{document}

%%%%%%%%% TITLE
\title{Batch-based Model Registration for Fast 3D Sherd Reconstruction}

\author{Jiepeng Wang$^1$ \and
Congyi Zhang$^1$ \and
Peng Wang$^1$ \and
Xin Li$^3$ \and
Peter J. Cobb$^1$ \and
Christian Theobalt$^2$ \and
Wenping Wang$^3$\thanks{Corresponding author.} \and
\\
$^1$The University of Hong Kong \qquad
$^2$Max Planck Institute for Informatics \qquad \\
$^3$Texas A\&M University
}

\maketitle
\ificcvfinal\thispagestyle{empty}\fi

\begin{abstract}
3D reconstruction techniques have widely been used for digital documentation of archaeological fragments. However, efficient digital capture of fragments remains as a challenge. 
In this work, we aim to develop a portable, high-throughput, and accurate reconstruction system for efficient digitization of fragments excavated in archaeological sites. 
To realize high-throughput digitization of large numbers of objects, an effective strategy is to perform scanning and reconstruction in batches. 
However, effective batch-based scanning and reconstruction face two key challenges: 1) how to correlate partial scans of the same object from multiple batch scans, and 2) how to register and reconstruct complete models from partial scans that exhibit only small overlaps. 
To tackle these two challenges, we develop a new batch-based matching algorithm that pairs the front and back sides of the fragments, and a new Bilateral Boundary ICP algorithm that can register partial scans sharing very narrow overlapping regions. 
Extensive validation in labs and testing in excavation sites demonstrate that these designs enable efficient batch-based scanning for fragments. 
We show that such a batch-based scanning and reconstruction pipeline can have immediate applications on digitizing sherds in archaeological excavations. Our project page: \url{https://jiepengwang.github.io/FIRES/}.

\end{abstract}

\section{Introduction}\label{sec:introduction}

Digital documentation of archaeological artifacts is widely used in archaeological heritage preservation and virtual restoration \cite{nemoto2023boatrestore,bea2017geometricdocu,hou2018novel, sizikova2017wall}. 
Sherds, also referred to as {\em fragments}, 
are the most commonly uncovered artifacts during archaeological excavations and they carry rich information about past human societies, so their 3D shapes need to be accurately reconstructed and digitally archived for analysis and preservation. 
At an excavation site, typically hundreds of fragments are uncovered in a day, far beyond the scanning and reconstruction capacity of existing imaging systems \cite{Porter2015SimpleRig,Ahmed2014DitializationStructuredLight,Magnani2014TwoSidesMerge,Karasik2008ScanMultiFragmentsWithFrame}. 
Hence, there is high demand for methods and systems capable of scanning and reconstructing hundreds of fragments per day. 

\begin{figure}
  \centering
  \mpage{0.435}{\includegraphics[width=\linewidth]{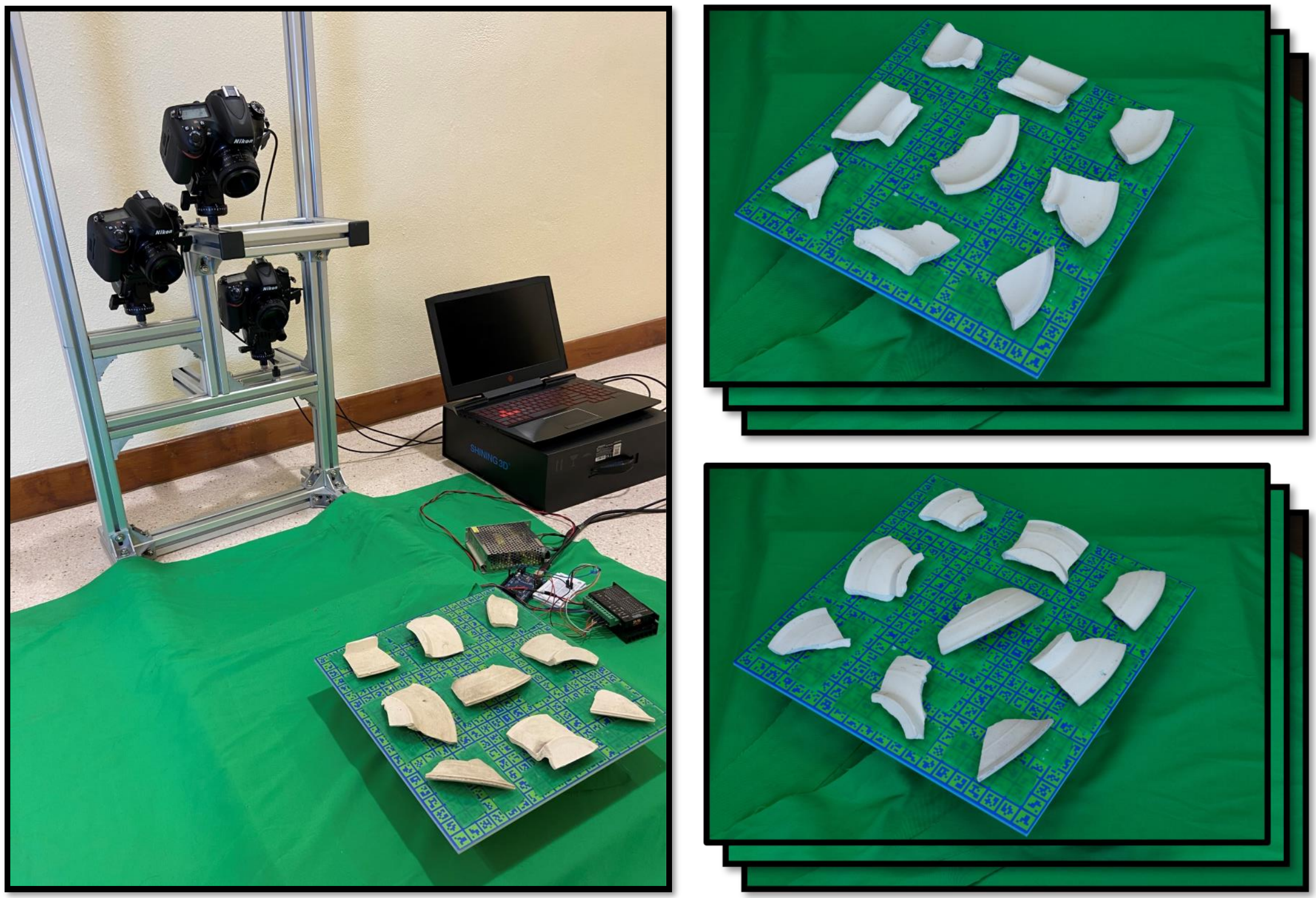}}
  \mpage{0.525}{\includegraphics[width=\linewidth]{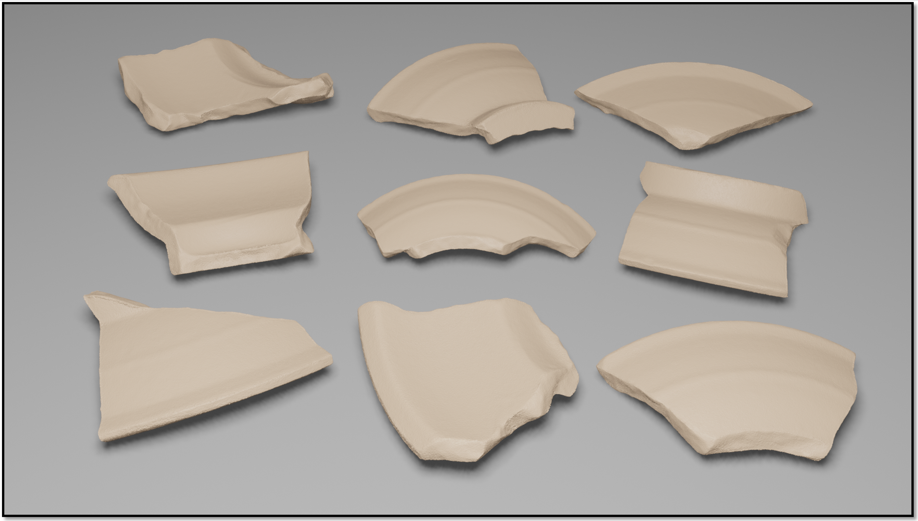}}

  \mpage{0.15}{(a)}
  \mpage{0.2}{(b)}
  \mpage{0.5}{(c)}
  
  \caption{{\bf Illustration of batch processing of sherds.} (a) Fast image acquisition device; 
  (b) Reference images of the front sides and back sides of a batch of fragments;
  and (c) Reconstructed 3D sherds. }
  \label{fig:teaser}
\end{figure}

The most promising approach to high throughput scanning of sherds is via batch processing. A straightforward flow of batching processing works as follows. First, all the fragments to be processed are divided into groups of 5 - 20 fragments per group,  with the specific number depending on the sizes of the fragments. For each group, the following three steps of operation are performed. 

{\em Step 1} - {\bf Front side reconstruction}. The group of fragments are manually placed flat on a turning-table and imaged by several fixed  cameras in a couple of minutes.  For easy of reference, we call the upward sides of the fragments at this moment their {\em front sides}. See {\bf Fig. \ref{fig:teaser} (a)}. Then the acquired images are used by an off-the-shelf MVS software to automatically reconstruct the 3D  models of the front sides. 

{\em Step 2} - {\bf Back side reconstruction}. In order to scan the complete models, the fragments are manually flipped over and placed again on the turntable. Then the back sides of the fragments are scanned ({\bf Fig. \ref{fig:teaser} (b)}) and reconstructed in a similar fashion.  

{\em Step 3} - {\bf Full model registration}.
Now we have obtained the reconstructed partial 3D models of the front sides and back sides of the group of fragments obtained in Step 1 and Step 2. 
Then, for each fragment, the two 3D models for its corresponding front side and back side which are also called the {\em front scan} and {\em back scan}, are automatically identified and registered together to produce the complete 3D model of the fragment (as shown in {\bf Fig. \ref{fig:teaser}(c)}). 

{\bf Challenges.} 
While steps 1 and 2 in the above are technically straightforward in principle, step 3 of full model registration poses new technical challenges that need to be tackled with novel techniques to achieve. The main challenges are two-fold. First, given the front scan of each fragment, it is nontrivial to efficiently and reliably find its corresponding back scan from the back side batch. This is a known outstanding problem {\bf \cite{fan2016automated}} with no effective solution. 
Second, once a pair of the front scan and back scan of the same fragment are identified, in order to apply the ICP registration method to register them together to obtain the full 3D model of the fragment, it is critical to develop robust techniques to provide reliable initial relative pose estimation for the ICP to start. Furthermore, we note that the front and back scans commonly have very small overlapping regions, i.e. the thin side strip regions of the fragment (See the figure below),\begin{wrapfigure}{r}{0.45\linewidth}
\vspace{-20pt}
  \begin{center}
    \includegraphics[width=\linewidth]{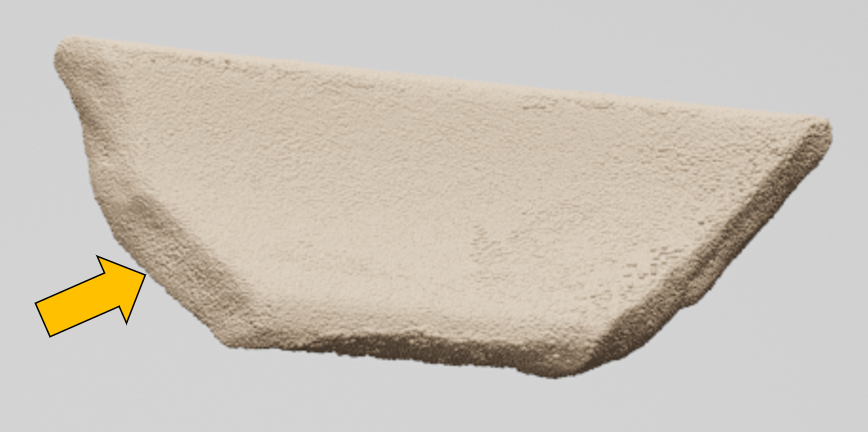}
  \end{center}
  % \vspace{-5pt}
    % \caption{Strip.}
    \label{fig:illus_strip}
    \vspace{-20pt}
\end{wrapfigure}which usually contain many repetitive fractured patterns and lack sufficient unique geometric features. Therefore, in the presence of such small overlap between the front and back scans, 
it presents significant challenges to existing ICP-like methods \cite{Ying2009ScaleICP,Chetverikov2002TrimmedICP,Yang2015GoICP} and feature-based registration methods \cite{Rusu2009FPFH,Mellado2014Super4PCS,Zhou2016FGR}. It is hard to ensure that points (or a feature) from one side used in registration have proper corresponding points (or a feature) on the other side.
Again, there are no existing registration methods that can reliably solves these problems to achieve successful registration to produce accurate full 3D models of the fragments in an automatic, fast, and robust manner. 

To address these issues, we provide a novel contour-based solution to solve batch matching and a boundary-aware ICP method to tackle fine registration to obtain complete models. 
We first discuss the problem of matching the front and back scans. We observe that 
the 3D contour of a fragment can be approximated by its 2D projection on the fitting
plane (e.g., using PCA), and the contour of this 2D projected region can be used as a reliable shape representation of the the fragments. Clearly, the 2D contours of the front scan and back scan of the same fragment should be highly similar. Hence, we devise a novel, intrinsic shape descriptor to encode each 2D contour, and then perform fast comparison of the shape descriptors of all the
2D contours to identify the pairing between the corresponding front and back scans. Our experiments show that this method is efficient and robust in finding all the matching pairs. 

After the pair of front and back scans are identified, we need to estimate a good initial alignment of the two scan in order to proceed with ICP registration. In this regards, we point out that, when a pair of scans are successfully matched, the above method based on 2D-contours also produces a good alignment of the 2D contours of the two scans, which induces a good initial alignment 3D scans themselves for the subsequent ICP iterations. The details of this step with explained in detail later. 

Finally, after each fragment's front and back scans are paired and initially aligned,  we will register them to get a complete 3D sherd model. To address of the issue of small overlap and the lack of salient textures and features in the small overlapping region of the two scans, we propose a \emph{boundary-based ICP} method to ensure robust ICP iterations. Specifically, we extract the 3D boundary points of one scan and iteratively search and minimize their distances to their corresponding points on the other side. Here the 3D boundary points refer to the points on the boundary of the point cloud surface of the front scan or the back scan. 
we use these 3D boundary points rather than all the points in the overlapping region for closest point searching because it is more likely for such boundary points to find their corresponding points on the other scan.  

Enabled by our novel batch-based model registration method, the partial front and back 3D scans be efficiently and automatically registered to produce complete 3D models of sherds in high throughput.  To validate our method, 
we built a turntable-based image acquisition device to capture the images of multiple fragments in a batch mode, as the input of our batch-based registration method.
We demonstrate that this batch-based approach can scan and reconstruct over 700 fragments in 10 working hours
, with high reconstruction accuracy, which is significantly faster than existing image acquisition systems and meets the throughput demand of archaeological field works. 

To summarize, we made the following contributions:

\begin{itemize}
    \item \textbf{Batch-based matching}: To support high-throughput batch-mode scanning and reconstruction of sherds, we propose a novel matching algorithm for matching the front partial 3D scans and the front partial 3D scans of a group of sherds. This solves an outstanding problem arising in batch-based sherd reconstruction. 
    
    \item \textbf{An improved registration method}: We develop an improved ICP method, called a {\em Bilateral Boundary ICP} (BBICP), that overcomes the issue of small overlaps to robustly and accurately register the corresponding front and back 3D scans of each fragment to produce its complete 3D model. 

    \item \textbf{Validation and dataset:} We conducted extensive validation to demonstrate that our algorithmic contributions enables efficient and robust scanning and reconstruction of sherd that is more than 10 times faster than existing 3D acquisition systems for sherds. We also a dataset containing 123 fragments of different geometries, sizes, and textures to  to facilitate future research on batch-based model registration for 3D reconstruction of archaeological fragments. 

\end{itemize}

The code and dataset will be made publicly available.  
\section{Related Work}

\begin{figure*}[!ht]
    \centering
    \includegraphics[width=\linewidth]{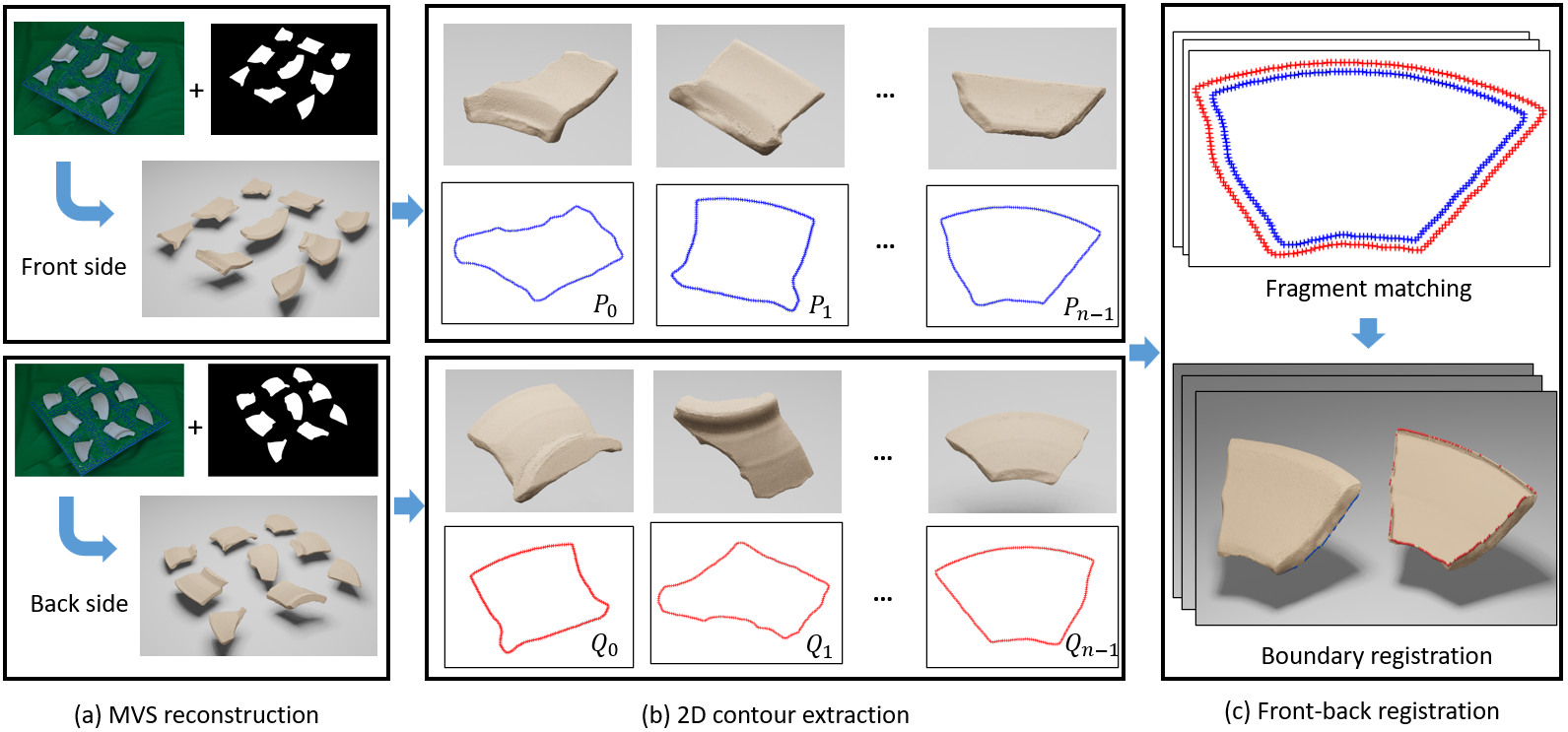}
    \caption{\textbf{Overview.} (a) Given the images of two sides, we first perform multi-view reconstruction with selected image regions of two sides with high efficiency.
    % (Section \ref{sec:img_acq}). 
    (b) With the reconstructed models, we then separate the models into individual fragments in each side and extract their maximum 2D contour (Section \ref{sec:front_back_matching}). (c) Based on these 2D contours, we propose a matching strategy to find correct matches of different fragments between their two sides. For each matching pair, local refinement is performed to get a complete and tight model (Section \ref{sec:registration_sicp})} 
    \label{fig:method_overview}
\end{figure*}

\subsection{Batch processing of sherds}\label{sec:related_works_sys}
Current data acquisition techniques can be generally categorized into 1) photogrammetry~\cite{Porter2015SimpleRig,Sapirstein2018Precision}, 2) structured light scanning~\cite{Ahmed2014DitializationStructuredLight,Karasik2008ScanMultiFragmentsWithFrame}, and 3) laser scanning~\cite{Magnani2014TwoSidesMerge} based systems. 
Because of their cost, portability, and ease of usage, photogrammetry devices are more widely adopted in  archaeology~\cite{Di2018FragmentsAnalysis}. In order to get a complete reconstruction, each object needs to be captured through multiple views. 
For example, the piece is often held to stand 
with the help of putty or an eraser. This setting, however, introduces tremendous manual labor and time during data acquisition.
In the data processing stage, the merge of different sides of the objects also needs a laborious trimming to remove extraneous material.
Users face the problems of frequent interactions and computer-processing time during the capturing and post-processing period to get complete models. 
Therefore, most existing approaches focus on processing a single object and cannot scale beyond a few dozen objects per day.

There are also hardware systems that combine multiple types of devices to provide hybrid solutions to record different aspects of a target object. 
However, to our best knowledge, all existing systems~\cite{Kasper2012IJRR,Singh2014ICRA,Hodan2017WACV,Kaskman2019ICCVW} in this category focus on reconstructing only visible region of objects on the turntable rather than getting complete 3D models. 
Also, most of them do not consider efficiency to be a key requirement, and hence, involve tremendous labor during acquisition or post processing. 

To realize a larger scale data acquisition, a practical approach is to process multiple fragments simultaneously. \cite{Karasik2008ScanMultiFragmentsWithFrame} used a specially designed frame to hold multiple fragments for faster scanning. 
But this system fails to get complete models because part of the fragments are occluded by the frame. \cite{fan2016automated} propose a 3D scanning system that can digitize fresco fragments by scanning then merging the the two sides of multiple pieces. 
However, the view planning process 
is time-consuming, making the acquisition less efficient. Another unsolved problem of this system, as the authors stated, is how to automatically find matches between the two partial 3D scans of the front side and back side of each fragment for registering them to get a complete reconstruction.

So far, global matching of two sides of pottery fragments in a batch-based data acquisition scheme remains an outstanding and challenging issue. We tackle this challenge by developing a novel contour-based matching strategy that enables scanning fragments laid flat on a turntable in batch mode. 

\subsection{Registration for 3D reconstruction}\label{sec:related_works_registration}

A 3D model registration method is needed for merging the front and back partial 3D scans to produce the complete 3D model of each fragment. A challenging issue here is that the these two partial scans only share a small overlap region, i.e. along the fractured strip surface. This poses a significant challenge to existing methods for reliable geometric registration.

Geometric registration~ \cite{Rosin2012RegSurvey} can be generally categorized into two types: global registration that finds a rough transformation between two surfaces, and local refinement that computes a precise transformation. Global registration methods are usually based on matches of local feature descriptors \cite{Zhou2016FGR,Rusu2009FPFH}, tuples of points \cite{Aiger20084PCS, Mellado2014Super4PCS}, or the branch-and-bound framework\cite{Yang2015GoICP}, while local refinement algorithms are often based on the iterative closest point (ICP) algorithm and its variants \cite{Chetverikov2002TrimmedICP,Ying2009ScaleICP,Pomerleau2013ICPVariants}.
Branch-and-bound based methods are often very expensive and prohibitive in handling very dense point sets.
Feature-based methods rely on salient texture or geometry features to find correct correspondence and transformations, which, unfortunately, are often unavailable on fragmented pieces. Thus, it's hard to apply current global registration methods to get a good initialization for subsequent ICP refinement. Tuples of points based methods and local refinement algorithms are sensitive to how much the two surfaces overlap with each other. However, between the front and back sides of each fragment, such an overlap is generally quite small.

ICP is also used in~\cite{Rusinkiewicz2008ScanMultiFragments} to align the front side to the back side of fresco fragments by assuming all the pieces are flat. This strategy is not suitable for sherds because sherd surfaces have curved shapes that are more complex than frescoes. Hence, it is hard to directly apply the existing ICP methods to robustly register the two partial scans of a sherd in our setting. 

In this work, we present a contour-based strategy that can provide a good initialization for registration, and a boundary-aware ICP method for robust fine registration of the front and back partial 3D scans to form a complete 3D sherd mode in the presence of small overlap. 

Recently, learning-based methods for point cloud registration~\cite{chen2019plade,wang2019dcp,choy2020dgr,zhang2020reg_overview_learning} have made remarkable progress. 
However, these learning-based methods often require a substantial amount of training data, which is the bottleneck of existing methods for fragment data acquisition. 
Our proposed batch-based model registration method generates large quantities of real partial fragment data and their reconstructed models.
Such data on fragments are currently lacking, and this hinders the development of effective fragment registration algorithms. 
Our work will provide such data and facilitate future research in applying learning-based techniques to fragment registration.

\section{Method}\label{Sec:Algo}

Our objective is to accurately and efficiently digitize a group of fragments into complete 3D models. 
To do this, we first capture multiple images that cover the front and back sides of the fragments in batches, through a custom-built device (See Fig. \ref{fig:teaser}(a)). Using these images, we can reconstruct a batch of front scans and a batch of back scans by UNet \cite{ronneberger2015u} to segment forground image regions and openMVS \cite{openMVS} for dense reconstruction (Fig. \ref{fig:method_overview}(a). 
Then, to register and reconstruct these models, we perform a two-step process:
(1) match the front and back scans from the two batches (Sec.~\ref{sec:front_back_matching}); (2) register the two sides into a complete 3D model (Sec.~\ref{sec:registration_sicp}).  
Both of these two steps present significant challenges and require substantial modifications and improvements to existing 3D matching and reconstruction technologies.

\subsection{Batch Matching to Pair Front and Back Sides}
~\label{sec:front_back_matching}

In order to register the partial scans of fragments in batches and reconstruct complete models, 
we first need to solve a \emph{pairing problem} that matches each fragment's front side with its respective back side from the two batches of partial scans. We refer this to as a \emph{front-back matching problem}. This problem is commonly encountered in group-based scanning systems, but reliable solutions to this problem remain an open research challenge~\cite{fan2016automated}.

A naive and immediate solution to this matching problem is to ensure that all fragments remain in (almost) the same positions when being flipped over, and then match the partial scans based on their locations. However, field tests revealed that this strategy is prone to errors. It is often challenging for the operator to guarantee that the positions of the fragments are not altered during flipping: sometimes, the shift is unintentional, while at other times, repositioning the pieces helps orient them to maximize visibility from the cameras. 
Therefore, position changes are hard to avoid, especially when dealing with multiple pieces in each group. This inconsistency can lead to mismatches and ultimately result in failed reconstruction.

We propose a batch matching scheme that (1) enables automatic and reliable pairing of corresponding partial scans, and (2) provides an initial alignment of the front and back sides. This is crucial for the subsequent registration task. 
Our approach is based on the observation that most curved fragments can be well approximated by their 2D projections onto fitting planes. 
Hence, we project the fragments onto their respective fitting planes, encode the resulting 2D contours with a shape descriptor, and then compare these descriptors to identify matching pairs. 

We first use PCA~\cite{Wold1987PCA} to project each partial scan onto its fitting plane (which passes through the point cloud centroid, with normal oriented along the smallest eigenvector). 
We then extract the 2D contour $\mathcal{C}$ of the projected point cloud using Alpha Shape~\cite{bernardini1997alphashape}, and sample $n_c$ ($n_c=200$) points uniformly clockwise along the contour, $\mathcal{C} = \{v_i\}_{i=0}^{n_c-1}$. 
\begin{wrapfigure}{r}{0.45\linewidth}
\vspace{-25pt}
  \begin{center}
    \includegraphics[width=\linewidth]{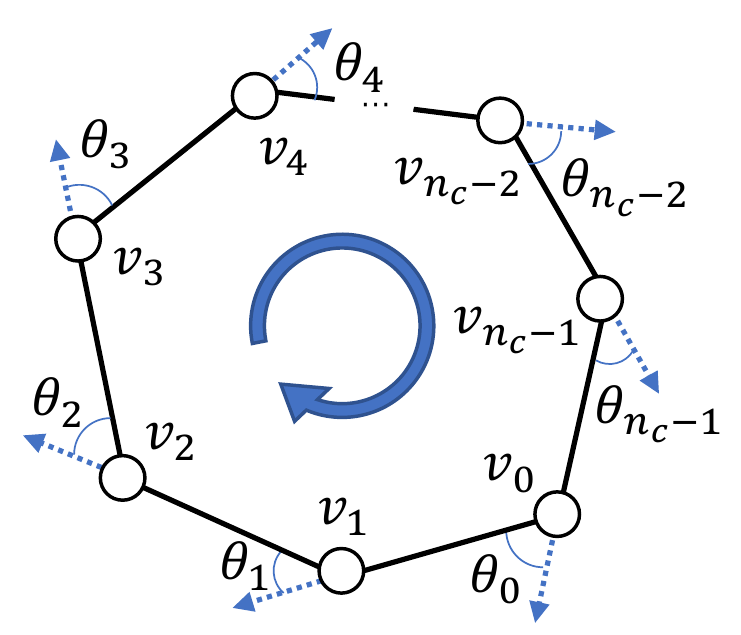}
  \end{center}
  \vspace{-5pt}
    %\caption{Contour description.}
    \label{fig:2d_contour_descriptor}
    \vspace{-10pt}
\end{wrapfigure}
With {\em turtle graphics}~\cite{Sederberg1993ShapeBlending}, $\mathcal{C}$ can be represented using the sequence of the turning angles $\{\theta \}_{i=0}^{n_c-1}$ (see the figure above). %\ref{fig:2d_contour_descriptor}
The sequence ${\theta}_{i=0}^{n_c-1}$, along with the common edge length $|\mathcal{C}|/n_c$, fully encodes the shape and size of the contour $\mathcal{C}$. 

To generate a shape descriptor for $\mathcal{C}$, we define a sequence of \emph{accumulative sums} of the turning angles, $\{\bar\theta_i \}_{i=0}^{n_c-1}$, where $\bar\theta_i = \sum_{j=0}^{i} \theta_j$, and form a vector $\Theta=\left( \bar \theta_i \right)^T \in \mathbb{R}^{n_c}$. 
This vector $\Theta$ serves as our {\em shape descriptor} of contour $\mathcal{C}$. 
Here using the accumulative sums of turning angles makes the descriptor more robust to noise and small shape perturbation than using turning angles themselves. 
The choice of starting vertex $v_0$ affects the descriptor. Hence, there are $n_c$ equivalent descriptors, each determined by the chosen starting vertex $v_0$. We denote these descriptors by $\Theta(\mathcal{C}, v_0)$ to emphasize this dependence.

\textbf{Batch matching.} 
Given two batches of 3D partial scans $\mathcal{P} =\{P_i\}_{i=0}^{n-1}$ and $\mathcal{Q}=\{Q_j\}_{j=0}^{n-1}$, respectively, from the front sides and back sides of a group of $n$ fragments, 
we first compute each partial scan's 2D projection and shape descriptors $\Theta_{P_i}$ and $\Theta_{Q_i}$. 
We can match the two sets, $\{P_i\}$ and $\{Q_j\}$, by matching descriptors. 
The distance between descriptors is defined using the $L_2$ norm: 
\begin{equation}
    E(\Theta_{P_i}, \Theta_{Q_j}) =  \min_{0\leq k \leq n_c-1} \left\|\Theta(P_i, v_0) - \Theta(Q_j, u_k)\right\|_2
    \label{eq:best_matching_local},
\end{equation}
where $v_0$ and $u_k$ are the first and $(k+1)$th sampled contour points from the projected $P_i$ and $Q_j$, respectively. 
With this batch matching scheme we can successfully pair one side of each partial scan $P_i$ with its corresponding other side $Q_j$ (see Fig.~\ref{fig:method_overview}(b)). And the found corner point pair $(v_0, u_k)$ indicates a good initial contour correspondence between $P_i$ and $Q_j$, which is important for the subsequent ICP-based registration.

\begin{figure}[ht]
  \centering

  \includegraphics[width=\linewidth]{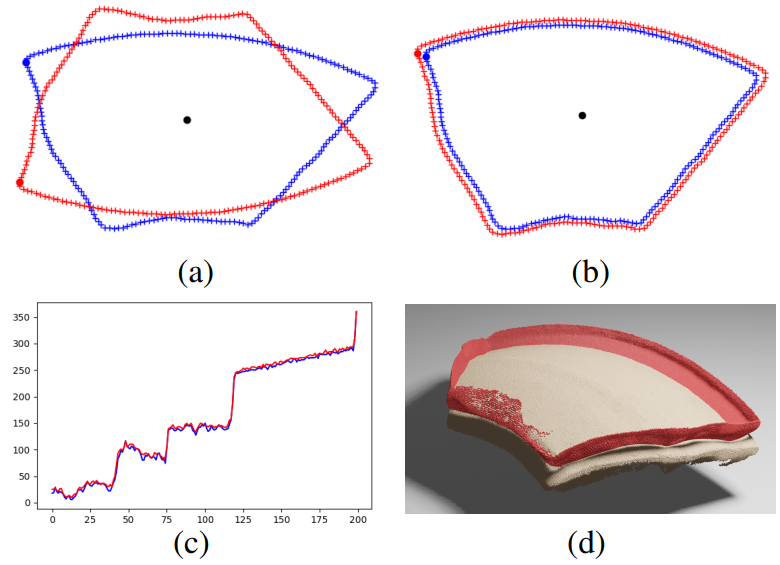}
  
  \caption{\textbf{Contour matching}. (a) 2D contours for the front side (blue) and back side (red); (b) Matching of the two 2D contours; (c) The plots of the shape descriptors of the two matched contours; and (d) The initial alignment of the two corresponding 3D partial scans as suggested by the 2D contour matching.} 
  % (the top scan is shown in red, the back scan is shown in yellow).}
  % Here, for better visualization, only the rim region of the front scan on the top is shown (in red) in order to avoid occluding the back scan (shown in yellow) at the bottom. }
  \label{fig:2d_contour_registration}
\end{figure}

\subsection{Registering the Front and Back Sides}
\label{sec:registration_sicp}

After each fragment's front and back sides are paired, we can then register them to produce a complete 3D sherd model. However, this registration a very challenging problem because the two sides of a fragment typically  have very little overlap and the fracture regions often lack geometric features. 
With the shape descriptor matching (in the previous step), a correspondence between the vertices of partial scans' contours has been established. Using this vertex correspondence, we align the centroids of the two 3D contours and then apply the algorithm of  \cite{Arun1987FitPointSets} to find an optimal rigid transformation to align the partial 3D scans $C_{P_i}$ and $C_{Q_j}$. This initial alignment serves as a good starting point for for the subsequent iterative registration process. 

\textbf{3D boundary extraction.}
Each reconstructed front or back side of a sherd is a point cloud surface of open-disk topology, and therefore, has a boundary that contacts, or is near, the holding board of the turntable at the bottom. 
To extract the boundary points of a partial 3D scan, we followed the strategy in \cite{linsen2001BoundaryExtraction}, but adopted the following procedure to improve efficiency. 
Given a reconstructed point cloud patch $P$, we identify boundary points by checking consistency from different views. First, a point $p$ in $P$ is projected to a pixel in input images where this point is visible. With the aid of masks generated for MVS reconstruction, we can determine whether a point in 3D is a candidate boundary point in one view. Specifically, we check the distance from $p$'s 2D projected point to the contour of the image mask. If this distance is smaller than a threshold in all the images in which the point is visible, we take this point as a candidate boundary point. 
In this way, we obtain a candidate set of boundary points, and remove non-boundary points to improve efficiency. 
The candidate set may still contain some outliers 
(i.e., some non-boundary points that are very close to the boundary contour). We then apply the widely adopted boundary extraction method \cite{linsen2001BoundaryExtraction} to this candidate set to extract the final set of boundary points. 

\textbf{Bilateral Boundary ICP (BBICP).} 
Next we use the extracted boundary points for model registration, i.e. computing a rigid transformation consisting of a rotation $R\in SO(3)$ and a translation $T \in \mathbb{R}^3$. 
Given two point clouds (front- and back-side partial models) $\mathcal{P}$ and $\mathcal{Q}$ as well as their boundary points $\mathcal{B_P}$ and $\mathcal{B_Q}$, the registration is performed by minimizing the sum of two terms: (1) one being the sum of the $L_2$ distances from the points in $\mathcal{B_P}$  to their corresponding closest points in $\mathcal{Q}$ , and (2) the other being the sum of the $L_2$ distances from the points in $\mathcal{B_Q}$ to their corresponding closest points in $\mathcal{P}$.

We formulate this problem as an ICP optimization problem based on the correspondences of the boundary points and the point sets. Specifically, for each point $b_{P_i}^k \in B_{P_i}$, we find its closest point in $Q_j$, and denote the found correspondences as $K_1=\{(b_{P_i},q_j)\}$. Similarly, for each point $b_{Q_j}^l \in B_{Q_j}$, we find its closest point in $P$, and denote the correspondences as $K_2=\{(p_i,b_{Q_j})\}$. We iteratively find the correspondences $K_1$ and $K_2$ and use the optimization method in \cite{Ying2009ScaleICP} to compute the final transformation by iteratively minimizing 
\begin{equation}
    \begin{array}{cl}
        \epsilon^k(K_1, K_2) = &
        \sum_{i=0}^{m-1}|| R^k b_i^P + T^k - q_i||^2  \\
        &+ \sum_{j=0}^{n-1}|| R^k p_j + T^k - b_j^Q||^2.
        \label{eq:objective_ICP}
    \end{array}
\end{equation}

\begin{figure}[ht]
  \centering
  \includegraphics[width=\linewidth]{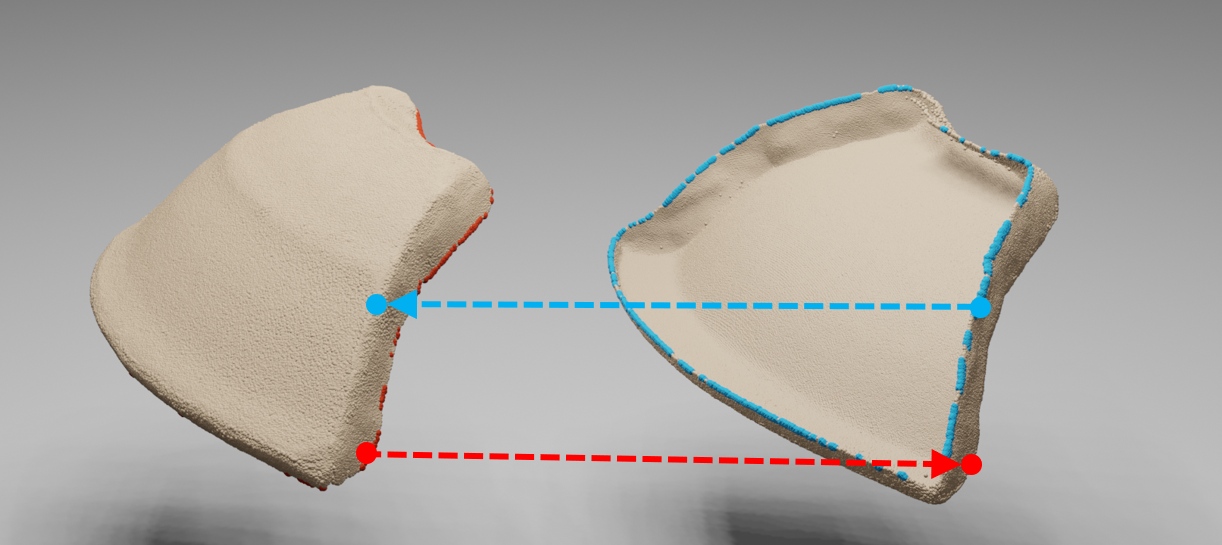}
  \caption{{\bf Illustration of BBICP.} 3D registration between the front side (whose boundary points are in blue) and  front side (whose boundary points are in red) in BBICP.}
  \label{fig:3d_boundary_registration}
\end{figure}

Fig. \ref{fig:3d_boundary_registration} illustrates an example of this \textbf{bilateral boundary} registration. The two sides of a fragment are registered with the help of their boundary points (in red and blue, respectively). During this process, only the boundary points of two sides are used to build correspondence with the other side. Compared with the existing ICP methods using all points to build correspondences for optimization, our method can ensure a large overlap and the convergence of ICP optimization.

\section{Experiments}\label{sec:exp}

We conducted a comprehensive quantitative evaluation of our designs and pipeline, which includes two aspects: reconstruction accuracy (Sec. \ref{sec:sys_accuracy}) and batch scanning efficiency (Sec. \ref{sec:sys_efficieny}). Since there is no public 3D fragment dataset available for systematic evaluation of batch capturing and reconstruction, we created a new dataset (Sec. \ref{sec:dataset}) and will release it to the public for comparisons. 
Furthermore, enabled by our model registration method, we built a prototype system and deployed it in archaeological excavations. The field experiments (Sec.~\ref{sec:field_experiments}) demonstrate the practicality and effectiveness of our method.

\subsection{Data Acquisition for Batch Scan Processing}
\label{sec:img_acq}

To validate our matching and registration algorithms, we build a batch-based image acquisition system, which is a customized hardware system (see Fig.~\ref{fig:teaser}) that consists of a turntable, three cameras mounted on an aluminum frame, and a controller module to synchronize the motion of turntable with the cameras shutters. To capture a \emph{group} of fragments in a batch mode, we place them flat on the turntable, and first take a set of pictures to capture their exposed sides, to be called the {\em front sides}.
Then the fragments are flipped manually on the turntable to photograph their {\em back sides}. We call all these pictures a \emph{batch}. And a total of 48 images are captured for each batch.  

Given a batch of captured images, we first segment the sherd regions in these images and generate their masks using UNet \cite{ronneberger2015u}. From these segmented regions, the partial 3D models (front or back side) of all the fragments on the table are then reconstructed using openMVS \cite{openMVS} efficiently. These result in multiple disjoint point clouds, each corresponding to one fragment (see some examples in Fig.~\ref{fig:method_overview} (a)), which will be used as the input of our model registration method.
Please refer to the supplementary for more details.

\subsection{Evaluation Dataset}
\label{sec:dataset}

We have built a dataset of ceramic fragments to evaluate/compare sherd acquisition and 3D reconstruction methods.  
The dataset consists of 123 fragments of varying shapes, sizes (2cm to 15cm in diameter), and thicknesses (0.3cm to 1cm), some of which are shown in the supplementary. These fragments were obtained by breaking several pottery items, whose original geometry before breaking was also scanned. We used a high-end EinScan Pro 2X \cite{EinScan} scanner in the fixed scan mode, with a reported accuracy of 0.04mm, to perform 3D scans. To capture all the surfaces of the fragments, we placed them on the scanner's turntable and captured 12 scans of each fragment's front side in a circular pattern. Then, we vertically clipped the fragments to scan their back sides. The partial scans were merged to obtain a complete 3D model using the scanner's software. We will publicly release this dataset, including images and 3D models, to facilitate comparative studies in fragment registration and reconstruction methods, as well as related research, such as 3D sherd reassembly and restoration.

\subsection{Reconstruction Accuracy}
\label{sec:sys_accuracy}

\begin{figure*}[h!t]
    \centering
    \includegraphics[width=\linewidth]{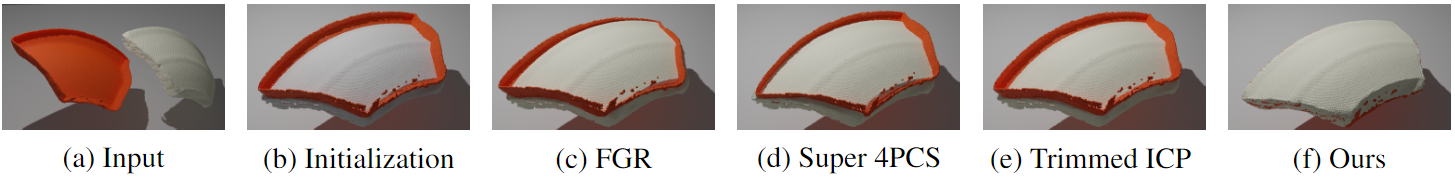}
     \caption{Comparison of different registration algorithms. (a) Input point cloud of the front and back partial 3D scans; (b) The initial alignment of the two sides. While our method is successful in this case, all the baseline methods fail to register the two scans, as there is obvious sliding error along the rim of the fragment. }
    \label{fig:exp_reg_comparsion}
\end{figure*}

To evaluate reconstruction accuracy, we adopted four widely used metrics: Accuracy (Accu), Completeness (Comp), Mean Absolute Error (MAE), and Error Standard Deviation (SD). We adopted their common definitions from the widely used Middlebury Benchmark~\cite{Seitz2006MVSBenchmark}. 
Definition details can be found in Section~5 of the supplementary file.
Given the reconstructed point cloud $R$ of a sherd, we align it with its corresponding scanned ground truth model $G$ in our dataset, using FGR~\cite{Zhou2016FGR} (for global alignment first) and ICP~\cite{besl1992icp} (for local refinement) .
The reconstruction accuracy is then measured by the difference between $R$ and $G$. 

\begin{table}[h!t]
   
    \caption{Accuracy evaluation of our fragment  reconstruction pipeline. The 123 fragments in our dataset are evaluated in 15 batchs. ID: batch ID; Num: the \emph{number} of fragments in a batch; Acc.(mm): average Accuracy of reconstructed fragments in a batch (smaller is better);  Comp.(\%): average Completeness of reconstructed fragments in a batch (higher is better); MAE(mm): average Mean Absolute Error of reconstructed fragments in a batch; SD(mm): average Standard Deviation of reconstructed fragments in a batch. \label{tab:pipeline_accuracy}}
    \centering
    % \begin{tabular}{cccccc}
    \begin{tabular}{>{\centering\arraybackslash}p{0.6cm} >{\centering\arraybackslash}p{0.8cm} >{\centering\arraybackslash}p{0.8cm} >{\centering\arraybackslash}p{0.8cm} >{\centering\arraybackslash}p{0.8cm} >{\centering\arraybackslash}p{0.8cm}}
    \toprule
    ID  & Num & Accu.$\downarrow $ \newline (mm) &  Comp.$\uparrow $  \newline (\%)   & MAE$\downarrow $ \newline (mm) & SD$\downarrow $ \newline (mm) \\
    \midrule
      1  & 9  & 0.14 & 98.5 & 0.08 & 0.06 \\ 
      2  & 9  & 0.14 & 98.39 & 0.08 & 0.06 \\ 
      3  & 7  & 0.15 & 93.74 & 0.09 & 0.06 \\ 
      4  & 9  & 0.15 & 97.27 & 0.08 & 0.07 \\
      5  & 8  & 0.11 & 98.47 & 0.07 & 0.06 \\
      6  & 8  & 0.12 & 97.64 & 0.07 & 0.05 \\ 
      7  & 8  & 0.13 & 96.80 & 0.07 & 0.07 \\ 
      8  & 9  & 0.19 & 95.65 & 0.10 & 0.07 \\ 
      9  & 9  & 0.19 & 94.42 & 0.10 & 0.08 \\ 
      10 & 7  & 0.13 & 98.53 & 0.08 & 0.06 \\ 
      11 & 7  & 0.16 & 90.88 & 0.09 & 0.07 \\ 
      12 & 4  & 0.20 & 90.75 & 0.10 & 0.10 \\ 
      13 & 4  & 0.16 & 95.92 & 0.09 & 0.11 \\ 
      14 & 7  & 0.15 & 98.12 & 0.09 & 0.09 \\ 
      15 & 18 & 0.15 & 94.95 & 0.08 & 0.06 \\ 
    \midrule
    \textbf{Mean} & \textbf{8.2}  & \textbf{0.15} &  \textbf{96.00} &  \textbf{0.09} &  \textbf{0.07} \\ 
    \bottomrule
    
    \end{tabular}
    % }
   
\end{table}

Table \ref{tab:pipeline_accuracy} reports the \textbf{reconstruction accuracy} of all the sherds in our dataset, captured in 15 batches. 
Our method achieved an average reconstruction \emph{accuracy} $T_a = 0.15mm$, \emph{completeness} $p_c = 96.00\%$, and mean absolute error $MAE=0.09mm$. 
Note that according to the survey provided by the Middlebury Benchmark, SOTA MVS algorithms 
reach \textbf{pixel level accuracy}, namely, $90\%$ reconstructed points have accuracy within about one pixel~\cite{furukawa2009accurate}.
As the average accuracy reported in Table~\ref{tab:pipeline_accuracy}, 
our reconstruction similarly reaches a \textbf{pixel-level accuracy}\footnote{ 
the average size of the fragments in our experiment is around $60mm$ wide, which occupies about $400$ pixels in the capture images. This means a pixel in the images can represent about $0.15mm$ (i.e., $0.15mm/pixel = 60mm/400pixels$).}. 
Note that these reconstruction errors include those from both MVS and registration steps. 
This indicates that our registration algorithm does not introduce additional significant errors.  

\textbf{BBICP vs other Registration Methods.} A key component of our proposed pipeline is the BBICP registration algorithm, which enables effective registration between partial scans with small overlap.  
We compared BBICP with the three widely used registration methods: 1) FGR \cite{Zhou2016FGR}, 2) Super 4PCS \cite{Mellado2014Super4PCS}, and 3) Trimmed ICP \cite{Chetverikov2002TrimmedICP}. By replacing BBICP with each of these registration methods in our reconstruction step, we can compare the reconstruction results.  As shown in Table \ref{tab:registration}, the performance of our registration method is significantly better than baseline methods.

\begin{table}[h!t]
    \centering
    \caption{Quantitative comparisons of registration. Acc.(mm): average Accuracy of all 123 fragments;  Comp.(\%): average Completeness of all fragments.}
    
    \begin{tabular}{cccc}
    \toprule
        Method       & Accu.$\downarrow $ (mm) &  Comp.$\uparrow $ (\%)  \\
    \midrule
    FGR \cite{Zhou2016FGR}       &   5.11      &   11.37   \\
    Super 4PCS \cite{Mellado2014Super4PCS}       &   4.84      &   10.03   \\
    Trimmed ICP\cite{Chetverikov2002TrimmedICP} &   4.71      &   18.69   \\ 
    {Ours}                                   &  \textbf{0.15} &   \textbf{96.00}   \\
    \bottomrule
    \end{tabular}
    
    \label{tab:registration}
\end{table}

In these experiments, the front-back batch matching process yields relatively good initial alignments between the partial scans of the front and back sides. 
However, the small overlapping regions and lack of distinct features in the fracture regions make it difficult for the baseline methods (FGR, Super 4PCS, and ICP) to build reliable correspondences and achieve accurate registration. 
In contrast, our method, BBICP, exhibits greater robustness in handling such cases. By constructing boundary-based correspondences, BBICP enables us to identify more overlap regions and achieve more precise registration. 
Fig.~\ref{fig:exp_reg_comparsion} shows an example fragment on which different registration methods were run and compared.

\begin{figure}
    \centering
    \includegraphics[width=0.9\linewidth]{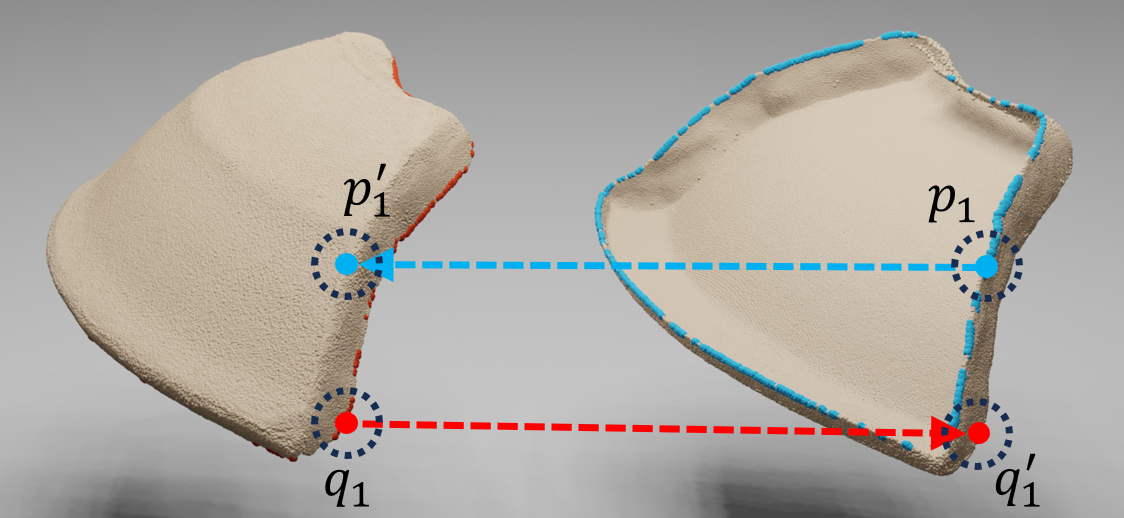}
    % \caption{2D illustration of FGR and Super 4PCS using filtered boundary points, which are marked as black dots.}
    \caption{Registering front and back scans of a fragment: each boundary point in one scan should find its correspondence among all the points in the other scan.}
    \label{fig:comp_bound}
    \vspace{-16pt}
\end{figure}

Note that in Table \ref{tab:registration}, the other methods used all points for registration. To align with the approach of our proposed method, we also conducted an experiment that uses only boundary points for registration. 
We conducted a quantitative evaluation on a batch containing 9 pieces. The errors for the FGR, Super-4PCS, and Trimmed ICP methods were $6.15 mm$, $3.87 mm$, and $4.41 mm$ respectively, whereas the error for our proposed method was only $0.13 mm$.
We can see that those methods do not perform well when only using boundary points. Fig.~\ref{fig:comp_bound} illustrates a typical scenario in which the feature-based method, such as FGR, fails to match corresponding partial scans using boundary points. 
Specifically, FGR first calculates FPFH for sampled points from their local neighboring regions, in order to construct local geometric features. Then, it establishes correspondences based on the similarity of these features.
Boundary points are derived from the red (front scan) and blue (back scan) contours. When considering the local neighboring regions of boundary points and their corresponding counterparts (e.g., points $p_1$, $q_1$ correspond to $p'_1$, $q'_1$), since each scan only covers one side of the fragment, boundary point $p_1$'s local neighboring region  contains only half geometric scan and is very different from $p'_1$'s neighborhood. 
Consequently, their computed local features are very different, resulting in incorrect correspondence computation. 
Similarly, Super4PCS struggles to establish accurate correspondences for boundary points due to low overlap and the lack of sufficiently distinct local geometries. Meanwhile, Trimmed ICP's performance is sensitive to the trimmed value.

\textbf{Archaeological Needs.} Within the archaeology community, the implications of automatic 3D scanning of fragments has not been well explored. The field is still developing requirements/standards on how accurately pieces should be scanned. 
Given that almost all prior work for documenting and measuring sherds has been undertaken manually, the introduction of digital methods represents a significant advancement, substantially exceeding the current state-of-the-art in terms of accuracy. 
The 3D models we have created are more than adequate for extracting 2D drawings for archaeological publications. 
Looking forward, our goal is to be as accurate as possible for the purposes of long-term archiving. We also plan to experiment with new analytical methods that are made possible by this new large-scale dataset, such as reassembling whole vessels back together. 
These future endeavors will help determine whether our current level of accuracy is sufficient for more detailed archaeological analyses or if further accuracy improvements are necessary.

\subsection{Efficiency}
\label{sec:sys_efficieny}

We compared the efficiency of our whole pipeline (including data acquisition and model reconstruction and registration) with several recent systems in literature. 
Table~\ref{tab:data_acquisition} lists different methods' estimated throughput within one hour. Our method is significantly faster than the other methods. Here, we also report the efficiency of preparation of GT models using EinScan (namely, E-GT in Table \ref{tab:data_acquisition}). Note that in our pipeline, most time is spent on the dense reconstruction step while our registration method is highly efficient. Because only a small subset (about 0.3$\%$) of source points distributed at boundaries require nearest-neighbor queries in each ICP iteration. It avoids building dense correspondences for all the points and significantly improve the registration efficiency.  
Please refer to Section 4 of the supplementary for a detailed explanation of how all these hourly throughput numbers were estimated. 
\begin{table}[h!t]
    \centering
    \caption{Comparison of data acquisition and reconstruction efficiency by different methods: the number of fragments that can be scanned within one work hour. E-GT denotes the throughput of preparation of GT data using EinScan.
    }
    
    \begin{tabular}{cccccccc}
    \toprule
  \cite{Rusinkiewicz2008ScanMultiFragments} &\cite{Brown2012Tools} &    \cite{fan2016automated} & \cite{Magnani2014TwoSidesMerge} & \cite{Porter2015SimpleRig} & \cite{Karasik2008ScanMultiFragmentsWithFrame}  & E-GT  & Ours \\           
    \midrule

    10 & 20 & 3 & 6 & 5 & 13  & 3 & 85  \\
    \bottomrule
    \end{tabular}
    
    \label{tab:data_acquisition}
\end{table}

\begin{figure*}[h!tb]
    \centering
    \includegraphics[width=\linewidth]{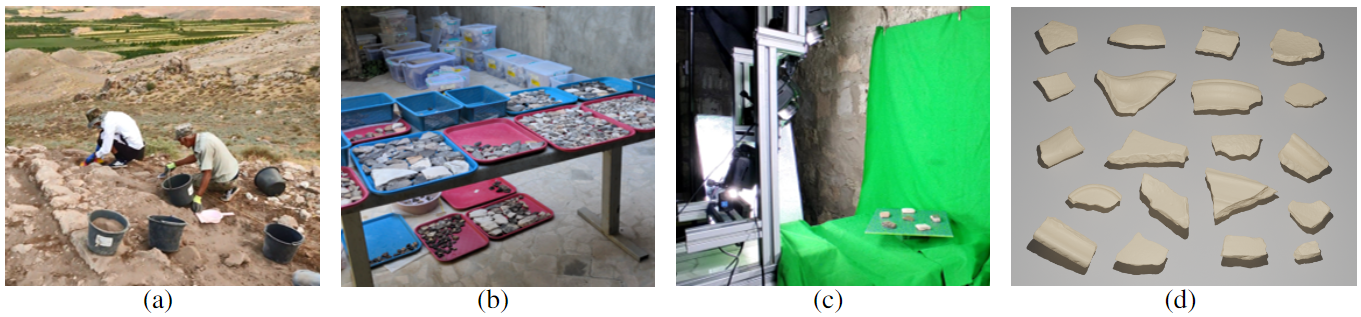}
    \caption{{\bf Prototype deployment.} (a) The excavation site; (b) The excavated fragments; (c) Our device deployed on the site; (d) Reconstructed 3D fragments.}
    \label{fig:deploy_armenia}
\end{figure*} 
\subsection{Field Validation and Application to Archaeological Excavation}%Experiments}
\label{sec:field_experiments}

We validated our algorithm by deploying our integrated pipeline and prototype system at an excavation site
in the summer of 2022 for two and half months. During this period, over 20,000 ceramic fragments were excavated, all 
digitized using our system. 
Fig.~\ref{fig:deploy_armenia} shows the excavation site (a), some excavated fragments (b), our device deployed in a local residence (c) which was adapted for use as a field lab, and some reconstructed models (d).  Our system achieved a throughput of about \textbf{730} fragments per day.

We quantitatively evaluated the reconstructed models from the site to test the field acquisition accuracy. Archaeologists randomly selected 26 fragments 
and scanned their corresponding ground-truth 3D models 
using EinScan \cite{EinScan}. 
The average reconstruction accuracy of these 26 fragments is about 0.16mm, which is similar to our lab experiments (0.15mm accuracy,  Table \ref{tab:pipeline_accuracy}). This validates our system's robustness in capturing accuracy when deployed in the field. 

Enabled by our model registration method, we provide a first \emph{practical solution to fast, accurate, and reliable digitization of a large number of sherds}. 
The successful large-scale experiment at the excavation site confirmed the feasibility and practicability of our batch scanning and reconstruction pipeline. 
We will make our code, data, and system configuration parameters publicly available and believe they can significantly alleviate the burden of archaeologists and boost the downstream applications, such as relic re-assembly. 

\subsection{Limitations}

Our batch matching method may fail when some fragments in a batch have very similar 2D contours or when some fragments are symmetrical (e.g., circular fragments), leading to ambiguities in the matching process. 
In such cases, during a practical batch scanning process, heuristics such as spatial locations of the fragments in the batch can be used to help resolve the matching ambiguity.

\section{Conclusions} % and Future work}
We proposed a novel batch-based model registration method, including a batch matching algorithm that matches partial 3D scans of the front and back sides of fragments and a new ICP-type method that registers the front and back sides sharing very narrow overlapping regions. Enabled by our method, we built a customized image acquisition device and established an automatic reconstruction pipeline that realize fast and precise reconstruction of fragments.
This system provides the first promising solution to meet the practical demand of archaeological fieldwork. And our system had been deployed and used at an excavation site in the summer of 2022. The field tests confirmed the feasibility and robustness of our method.

\section*{Acknowledgments}
We would like to thank all the people who have provided helpful suggestions and feedback for this project, including Lei Yang, Jiaming Xie, Yuhan Ping, Weilin Wan, Ruixing Jia, Yuan Liu and  Cheng Lin. This work was in part funded by a grant from the Research Grants Council (RGC) of the Hong Kong Special Administrative Region (HKSAR), China (project no. HKU 27602920).

\bibliographystyle{ieee_fullname}
\bibliography{paper_main.bbl}

\begin{thebibliography}{10}\itemsep=-1pt

\bibitem{Ahmed2014DitializationStructuredLight}
Namir Ahmed, Michael Carter, and Neal Ferris.
\newblock Sustainable archaeology through progressive assembly 3d digitization.
\newblock {\em World Archaeology}, 46(1):137--154, 2014.

\bibitem{Aiger20084PCS}
Dror Aiger, Niloy~J Mitra, and Daniel Cohen-Or.
\newblock 4-points congruent sets for robust pairwise surface registration.
\newblock In {\em ACM SIGGRAPH 2008 papers}, pages 1--10. 2008.

\bibitem{Arun1987FitPointSets}
K~Somani Arun, Thomas~S Huang, and Steven~D Blostein.
\newblock Least-squares fitting of two 3-d point sets.
\newblock {\em IEEE Transactions on pattern analysis and machine intelligence}, (5):698--700, 1987.

\bibitem{bea2017geometricdocu}
Manuel Bea and Jorge Ang{\'a}s.
\newblock Geometric documentation and virtual restoration of the rock art removed in arag{\'o}n (spain).
\newblock {\em Journal of Archaeological Science: Reports}, 11:159--168, 2017.

\bibitem{bernardini1997alphashape}
Fausto Bernardini and Chandrajit~L Bajaj.
\newblock Sampling and reconstructing manifolds using alpha-shapes.
\newblock 1997.

\bibitem{besl1992icp}
Paul~J Besl and Neil~D McKay.
\newblock A method for registration of 3-d shapes.
\newblock In {\em Sensor fusion IV: control paradigms and data structures}, volume 1611, pages 586--606. Spie, 1992.

\bibitem{Brown2012Tools}
Benedict Brown, Lara Laken, Philip Dutr{\'e}, Luc Van~Gool, Szymon Rusinkiewicz, and Tim Weyrich.
\newblock Tools for virtual reassembly of fresco fragments.
\newblock {\em International journal of heritage in the digital era}, 1(2):313--329, 2012.

\bibitem{Rusinkiewicz2008ScanMultiFragments}
Benedict~J Brown, Corey Toler-Franklin, Diego Nehab, Michael Burns, David Dobkin, Andreas Vlachopoulos, Christos Doumas, Szymon Rusinkiewicz, and Tim Weyrich.
\newblock A system for high-volume acquisition and matching of fresco fragments: Reassembling theran wall paintings.
\newblock {\em ACM transactions on graphics (TOG)}, 27(3):1--9, 2008.

\bibitem{openMVS}
Dan Cernea.
\newblock {OpenMVS}: Multi-view stereo reconstruction library.
\newblock 2020.

\bibitem{chen2019plade}
Songlin Chen, Liangliang Nan, Renbo Xia, Jibin Zhao, and Peter Wonka.
\newblock Plade: A plane-based descriptor for point cloud registration with small overlap.
\newblock {\em IEEE Transactions on Geoscience and Remote Sensing}, 58(4):2530--2540, 2019.

\bibitem{Chetverikov2002TrimmedICP}
Dmitry Chetverikov, Dmitry Svirko, Dmitry Stepanov, and Pavel Krsek.
\newblock The trimmed iterative closest point algorithm.
\newblock In {\em Object recognition supported by user interaction for service robots}, volume~3, pages 545--548. IEEE, 2002.

\bibitem{choy2020dgr}
Christopher Choy, Wei Dong, and Vladlen Koltun.
\newblock Deep global registration.
\newblock In {\em Proceedings of the IEEE/CVF conference on computer vision and pattern recognition}, pages 2514--2523, 2020.

\bibitem{Di2018FragmentsAnalysis}
L Di~Angelo, P Di~Stefano, and C Pane.
\newblock An automatic method for pottery fragments analysis.
\newblock {\em Measurement}, 128:138--148, 2018.

\bibitem{EinScan}
Einscan pro 2x.
\newblock \url{https://www.einscan.com/handheld-3d-scanner/einscan-pro-2x/}, 2020.

\bibitem{fan2016automated}
Xinyi Fan, Linguang Zhang, Benedict Brown, and Szymon Rusinkiewicz.
\newblock Automated view and path planning for scalable multi-object 3d scanning.
\newblock {\em ACM Transactions on Graphics (TOG)}, 35(6):1--13, 2016.

\bibitem{furukawa2009accurate}
Yasutaka Furukawa and Jean Ponce.
\newblock Accurate, dense, and robust multiview stereopsis.
\newblock {\em IEEE transactions on pattern analysis and machine intelligence}, 32(8):1362--1376, 2009.

\bibitem{Hodan2017WACV}
T. {Hodan}, P. {Haluza}, Š. {Obdržálek}, J. {Matas}, M. {Lourakis}, and X. {Zabulis}.
\newblock T-less: An rgb-d dataset for 6d pose estimation of texture-less objects.
\newblock In {\em 2017 IEEE Winter Conference on Applications of Computer Vision (WACV)}, pages 880--888, 2017.

\bibitem{hou2018novel}
Miaole Hou, Su Yang, Yungang Hu, Yuhua Wu, Lili Jiang, Sizhong Zhao, and Putong Wei.
\newblock Novel method for virtual restoration of cultural relics with complex geometric structure based on multiscale spatial geometry.
\newblock {\em ISPRS International Journal of Geo-Information}, 7(9):353, 2018.

\bibitem{Karasik2008ScanMultiFragmentsWithFrame}
Avshalom Karasik and Uzy Smilansky.
\newblock 3d scanning technology as a standard archaeological tool for pottery analysis: practice and theory.
\newblock {\em Journal of Archaeological Science}, 35(5):1148--1168, 2008.

\bibitem{Kaskman2019ICCVW}
R. {Kaskman}, S. {Zakharov}, I. {Shugurov}, and S. {Ilic}.
\newblock Homebreweddb: Rgb-d dataset for 6d pose estimation of 3d objects.
\newblock In {\em 2019 IEEE/CVF International Conference on Computer Vision Workshop (ICCVW)}, pages 2767--2776, 2019.

\bibitem{Kasper2012IJRR}
Alexander Kasper, Zhixing Xue, and Rüdiger Dillmann.
\newblock The kit object models database: An object model database for object recognition, localization and manipulation in service robotics.
\newblock {\em The International Journal of Robotics Research}, 31(8):927--934, 2012.

\bibitem{linsen2001BoundaryExtraction}
Lars Linsen.
\newblock {\em Point cloud representation}.
\newblock Univ., Fak. f{\"u}r Informatik, Bibliothek Technical Report, Faculty of Computer~…, 2001.

\bibitem{Magnani2014TwoSidesMerge}
Matthew Magnani.
\newblock Three-dimensional alternatives to lithic illustration.
\newblock {\em Advances in Archaeological Practice}, 2(4):285--297, 2014.

\bibitem{Mellado2014Super4PCS}
Nicolas Mellado, Dror Aiger, and Niloy~J Mitra.
\newblock Super 4pcs fast global pointcloud registration via smart indexing.
\newblock In {\em Computer Graphics Forum}, volume~33, pages 205--215. Wiley Online Library, 2014.

\bibitem{nemoto2023boatrestore}
Takashi Nemoto, Tetsuya Kobayashi, Masataka Kagesawa, Takeshi Oishi, Hiromasa Kurokochi, Sakuji Yoshimura, Eissa Zidan, and Mamdouh Taha.
\newblock Virtual restoration of ancient wooden ships through non-rigid 3d shape assembly with ruled-surface ffd.
\newblock {\em International Journal of Computer Vision}, pages 1--15, 2023.

\bibitem{Pomerleau2013ICPVariants}
Fran{\c{c}}ois Pomerleau, Francis Colas, Roland Siegwart, and St{\'e}phane Magnenat.
\newblock Comparing icp variants on real-world data sets.
\newblock {\em Autonomous Robots}, 34(3):133--148, 2013.

\bibitem{Porter2015SimpleRig}
Samantha~T Porter, Morgan Roussel, and Marie Soressi.
\newblock A simple photogrammetry rig for the reliable creation of 3d artifact models in the field: lithic examples from the early upper paleolithic sequence of les cott{\'e}s (france).
\newblock 2015.

\bibitem{ronneberger2015u}
Olaf Ronneberger, Philipp Fischer, and Thomas Brox.
\newblock U-net: Convolutional networks for biomedical image segmentation.
\newblock In {\em International Conference on Medical image computing and computer-assisted intervention}, pages 234--241. Springer, 2015.

\bibitem{Rusu2009FPFH}
Radu~Bogdan Rusu, Nico Blodow, and Michael Beetz.
\newblock Fast point feature histograms (fpfh) for 3d registration.
\newblock In {\em 2009 IEEE international conference on robotics and automation}, pages 3212--3217. IEEE, 2009.

\bibitem{Sapirstein2018Precision}
Philip Sapirstein.
\newblock A high-precision photogrammetric recording system for small artifacts.
\newblock {\em Journal of Cultural Heritage}, 31:33--45, 2018.

\bibitem{Sederberg1993ShapeBlending}
Thomas~W Sederberg, Peisheng Gao, Guojin Wang, and Hong Mu.
\newblock 2-d shape blending: an intrinsic solution to the vertex path problem.
\newblock In {\em Proceedings of the 20th annual conference on Computer graphics and interactive techniques}, pages 15--18, 1993.

\bibitem{Seitz2006MVSBenchmark}
Steven~M Seitz, Brian Curless, James Diebel, Daniel Scharstein, and Richard Szeliski.
\newblock A comparison and evaluation of multi-view stereo reconstruction algorithms.
\newblock In {\em 2006 IEEE computer society conference on computer vision and pattern recognition (CVPR'06)}, volume~1, pages 519--528. IEEE, 2006.

\bibitem{Singh2014ICRA}
A. {Singh}, J. {Sha}, K.~S. {Narayan}, T. {Achim}, and P. {Abbeel}.
\newblock Bigbird: A large-scale 3d database of object instances.
\newblock In {\em 2014 IEEE International Conference on Robotics and Automation (ICRA)}, pages 509--516, 2014.

\bibitem{sizikova2017wall}
Elena Sizikova and Thomas Funkhouser.
\newblock Wall painting reconstruction using a genetic algorithm.
\newblock {\em Journal on Computing and Cultural Heritage (JOCCH)}, 11(1):1--17, 2017.

\bibitem{Rosin2012RegSurvey}
Gary~KL Tam, Zhi-Quan Cheng, Yu-Kun Lai, Frank~C Langbein, Yonghuai Liu, David Marshall, Ralph~R Martin, Xian-Fang Sun, and Paul~L Rosin.
\newblock Registration of 3d point clouds and meshes: A survey from rigid to nonrigid.
\newblock {\em IEEE transactions on visualization and computer graphics}, 19(7):1199--1217, 2012.

\bibitem{wang2019dcp}
Yue Wang and Justin~M Solomon.
\newblock Deep closest point: Learning representations for point cloud registration.
\newblock In {\em Proceedings of the IEEE/CVF international conference on computer vision}, pages 3523--3532, 2019.

\bibitem{Wold1987PCA}
Svante Wold, Kim Esbensen, and Paul Geladi.
\newblock Principal component analysis.
\newblock {\em Chemometrics and intelligent laboratory systems}, 2(1-3):37--52, 1987.

\bibitem{Yang2015GoICP}
Jiaolong Yang, Hongdong Li, Dylan Campbell, and Yunde Jia.
\newblock Go-icp: A globally optimal solution to 3d icp point-set registration.
\newblock {\em IEEE transactions on pattern analysis and machine intelligence}, 38(11):2241--2254, 2015.

\bibitem{Ying2009ScaleICP}
Shihui Ying, Jigen Peng, Shaoyi Du, and Hong Qiao.
\newblock A scale stretch method based on icp for 3d data registration.
\newblock {\em IEEE Transactions on Automation Science and Engineering}, 6(3):559--565, 2009.

\bibitem{zhang2020reg_overview_learning}
Zhiyuan Zhang, Yuchao Dai, and Jiadai Sun.
\newblock Deep learning based point cloud registration: an overview.
\newblock {\em Virtual Reality \& Intelligent Hardware}, 2(3):222--246, 2020.

\bibitem{Zhou2016FGR}
Qian-Yi Zhou, Jaesik Park, and Vladlen Koltun.
\newblock Fast global registration.
\newblock In {\em European Conference on Computer Vision}, pages 766--782. Springer, 2016.

\end{thebibliography}

\end{document}